\documentclass[sn-basic]{sn-jnl}

\usepackage{graphicx}%
\usepackage{amsmath,amssymb,amsfonts}%
\usepackage{amsthm}%
\usepackage[title]{appendix}%
\usepackage{listings}%
\usepackage{bm}
\usepackage{mathdots}
\usepackage{array} %
\newcommand{\scdots}{\cdot\!\cdot\!\cdot}
\usepackage{caption}
\usepackage{url}
\begin{document}

\title[NMF-LAB]{Applying non-negative matrix factorization with covariates to label matrix for classification}

\author[1]{\fnm{Kenichi} \sur{Satoh}}

\email{kenichi-satoh@biwako.shiga-u.ac.jp}

\affil[1]{\orgdiv{Faculty of Data Science}, \orgname{Shiga University}, \orgaddress{\street{Banba 1-1-1}, \city{Hikone}, \postcode{522-8522}, \state{Shiga}, \country{Japan}}}


\abstract{
Non-negative matrix factorization (NMF) is widely used for dimensionality reduction and interpretable analysis, but standard formulations are unsupervised and cannot directly exploit class labels. Existing supervised or semi-supervised extensions usually incorporate labels only via penalties or graph constraints, still requiring an external classifier. We propose \textit{NMF-LAB} (Non-negative Matrix Factorization for Label Matrix), which redefines classification as the inverse problem of non-negative matrix tri-factorization (tri-NMF). Unlike joint NMF methods, which reconstruct both features and labels, NMF-LAB directly factorizes the label matrix $Y$ as the observation, while covariates $A$ are treated as given explanatory variables. This yields a direct probabilistic mapping from covariates to labels, distinguishing our method from label-matrix factorization approaches that mainly model label correlations or impute missing labels. Our inversion offers two key advantages: (i) class-membership probabilities are obtained directly from the factorization without a separate classifier, and (ii) covariates, including kernel-based similarities, can be seamlessly integrated to generalize predictions to unseen samples. In addition, unlabeled data can be encoded as uniform distributions, supporting semi-supervised learning. Experiments on diverse datasets, from small-scale benchmarks to the large-scale MNIST dataset, demonstrate that NMF-LAB achieves competitive predictive accuracy, robustness to noisy or incomplete labels, and scalability to high-dimensional problems, while preserving interpretability. By unifying regression and classification within the tri-NMF framework, NMF-LAB provides a novel, probabilistic, and scalable approach to modern classification tasks.}


\maketitle

\section{Introduction}\label{sec1}

Non-negative matrix factorization (NMF) is a widely used method for dimensionality reduction, feature extraction, and interpretable analysis in high-dimensional non-negative datasets \citep{lee1999,cichocki2009,gillis2014}. By decomposing an observation matrix into non-negative factors, NMF yields a parts-based representation that, when column-normalized, allows a probabilistic interpretation and naturally supports soft clustering.

\textit{Basic Model.} Formally, the basic model approximates an observation matrix $Y \in \mathbb{R}_+^{P \times N}$ as
\begin{equation}
Y \approx X B ,
\end{equation}
where $X \in \mathbb{R}_+^{P \times Q}$ and $B \in \mathbb{R}_+^{Q \times N}$ are constrained to be non-negative. This formulation is the foundation for numerous extensions that incorporate supervision or side information.

Classical NMF is unsupervised and does not incorporate class labels. To improve discriminative power, many supervised or semi-supervised extensions have been proposed. Supervised NMF (SNMF) and robust semi-supervised NMF (RSSNMF) incorporate labels indirectly through additional penalty terms or constraints so that samples with the same label share similar representations \citep{leuschner2018,wang2015}. Semi-supervised NMF (SSNMF) and joint NMF (jNMF) frameworks go further by jointly factorizing both the feature matrix and the label matrix $Y$ under a shared latent code \citep{wu2015,haddock2021}. For example, \citet{haddock2021} proposed SSNMF models that simultaneously provide a topic model and a classification model, thereby integrating representation learning and prediction within the same factorization framework. Another line of work is low-rank label-matrix factorization for multi-label learning, which aims to exploit label correlations or to handle missing labels. For example, \citet{yu2014} developed scalable methods for large-scale multi-label classification with incomplete label assignments, while \citet{zhang2015} proposed to construct label-specific features to improve discrimination. These approaches mainly focus on modeling label dependencies or imputing missing labels, but they do not provide a direct covariate-to-label mapping within the factorization.

In contrast, we propose a new framework, \textit{NMF-LAB} (Non-negative Matrix Factorization for Label Matrix), which directly addresses this predictive task by inverting tri-NMF with covariates, treating the label matrix as the observation matrix to be reconstructed from covariates. This architectural choice distinguishes our framework from joint factorization methods like SSNMF. While SSNMF seeks to simultaneously reconstruct both the feature matrix $A$ and the label matrix $Y$ through a shared latent space $S$ (i.e., $\min \|A-X_{A}S\|^{2}+\lambda\|Y-X_{Y}S\|^{2}$), our approach does not model the feature space at all. NMF-LAB is a pure predictive (or discriminative) model focused solely on the mapping from $A$ to $Y$ (i.e., $\min \|Y-X\Theta A\|^{2}$), which is the key to enabling direct probabilistic estimation without an external classifier.

Graph-regularized NMF has also been proposed to exploit the manifold structure of the feature space $X$ through Laplacian or similarity constraints \citep{cai2011,zhang2020}. Our use of kernel-based covariates differs fundamentally: the kernel operates directly on $A$ (the covariate space), inducing local averaging among similar individuals and thereby generalizing the mapping $A \mapsto Y$ to unseen samples. Other variants such as Task-driven NMF \citep{bisot2016} or discriminant/projective NMF \citep{yang2010,guan2013} incorporate labels through auxiliary penalties or projection constraints to enhance discriminability. In all such cases, however, labels act only as side information. By contrast, the idea of treating the label matrix itself as the observation matrix $Y$, while simultaneously using covariates $A$ to generalize predictions, has been relatively unexplored. NMF-LAB addresses this gap by employing tri-factorization ($Y \approx X\Theta A$; \citealp{ding2006}) and casting classification as its inverse problem. Table~\ref{tab:checklist} compares representative 
NMF-based approaches with our proposed framework NMF-LAB. Most supervised or semi-supervised NMF variants incorporate 
labels only indirectly (e.g., via auxiliary constraints or regularization). In contrast, NMF-LAB directly factorizes the label matrix as the 
observation and leverages covariates to generalize predictions. In the table, a blank entry indicates that the corresponding 
information is not used, while a check mark indicates that it 
is used or required.
\begin{table}[t]
\captionsetup{width=\textwidth}
\centering
\caption{Comparison of existing NMF-related approaches and the proposed NMF-LAB. 
A check mark in the column “Labels” indicates that class-label information is used 
during training (either as constraints, penalties, or direct factorization). “Classifier” denotes whether an additional external classifier is required for prediction, 
and “Cov$\to$Lab” indicates whether the method provides a covariate-to-label mapping 
within the factorization}
\label{tab:checklist}
\begin{tabular}{|p{4cm}|c|c|c|}
\hline
Approach & Labels & Classifier & Cov$\to$Lab \\
\hline
Classical NMF &  & \checkmark &  \\\hline
SNMF / RSSNMF & \checkmark & \checkmark &  \\\hline
SSNMF / jNMF & \checkmark &  &  \\\hline
Low-rank label-matrix fact. & \checkmark &  &  \\\hline
Graph-regularized NMF &  & \checkmark &  \\\hline
Task-driven / Discriminant NMF & \checkmark & \checkmark &  \\\hline
\textit{NMF-LAB (proposed)} & \checkmark &  & \checkmark \\\hline
\end{tabular}
\end{table}

In particular, our framework highlights the \textit{forward\textendash inverse duality}
 of tri-NMF: whereas previous studies employed covariates to explain or predict outcomes in the standard (\textit{forward}) formulation \citep{satoh2023,satoh2024,satoh2025}, NMF-LAB inverts this relationship and treats the label matrix itself as the observation to be reconstructed (\textit{inverse problem}). This inversion not only provides conceptual novelty but also offers practical flexibility, establishing a unified perspective that encompasses both regression and classification tasks. Crucially, this single framework gives rise to two distinct variants: a linear model (\textit{NMF-LAB (direct)}) that offers strong feature-level interpretability, and a non-linear model (\textit{NMF-LAB (kernel)}) that achieves high predictive accuracy comparable to state-of-the-art methods. The fundamental trade-off between interpretability and predictive power, explored through these two variants, constitutes one of the central findings of this work. For clarity, throughout the remainder of this paper we use $P$ 
to denote the number of classes when $Y$ represents a label matrix.

In summary, \textit{NMF-LAB} unifies feature learning and classification within a single regression-like formulation. Its key contributions are:
\begin{itemize}
\item \textit{Inverse problem formulation}: casting classification as the inverse problem of tri-NMF by interchanging the roles of $Y$ and $A$, thereby extending the growth-curve view of NMF with covariates.
\item \textit{Direct probability estimation}: obtaining class-membership probabilities directly from the factorization without requiring an external classifier.
\item \textit{Interpretability and partial identifiability}: each basis 
  vector often corresponds closely to a class, consistent with recent 
  results on partial identifiability of NMF under mild conditions 
  \citep{gillis2023}.
\item \textit{Robustness to label noise}: smoothing out spurious errors through factorization, leading to resilience against mislabeled data.
\item \textit{Semi-supervised flexibility}: encoding unlabeled samples as uniform or prior-weighted distributions, which enables a natural extension to semi-supervised settings.
\end{itemize}

The rest of the paper is organized as follows. 
Section~2 reviews NMF with covariates in the standard forward problem formulation.
Section~3 introduces NMF-LAB as the inverse problem formulation.
Section~4 presents methodological details and extensions, including the use of kernel-based covariates.
Section~5 reports empirical evaluations, focusing on predictive accuracy, interpretability, robustness to label noise, semi-supervised settings, and scalability.
Section~6 concludes with a summary and future directions.

\section{NMF with covariates: formulation and forward problem}\label{sec2}

Let the observation matrix be $Y=(\bm{y}_1,\dots,\bm{y}_N)=(y_{p,n})_{P\times N}$, 
consisting of $P$ variables measured on $N$ individuals. For a chosen rank $Q \le \min(P, N)$, we approximate $Y$ by the product of  
a basis matrix $X=(x_{p,q})_{P \times Q}$,  
a parameter matrix $\Theta=(\theta_{q,k})_{Q \times R}$,  
and a known covariate matrix $A=(a_{k,n})_{R\times N}$:
\begin{equation}
\mathop{Y}_{P \times N} \approx \mathop{X}_{P \times Q}\,\mathop{\Theta}_{Q \times R}\,\mathop{A}_{R\times N},
\label{XCA}
\end{equation}
where all elements of $Y$, $X$, $\Theta$, and $A$ are non-negative. Each column of $X$, denoted $\bm{x}_q$, represents a latent factor that additively contributes to the observed data. For individual $n$, (\ref{XCA}) reduces to
\begin{equation}
\bm{y}_n \approx X \Theta \bm{a}_n,
\qquad n=1,\ldots,N. \label{XCa}
\end{equation}

The mean structure in (\ref{XCA}) coincides with the classical growth curve model 
(GCM) of \citet{potthoff1964}. In GCM, the basis matrix $X$ is typically 
specified \emph{a priori} (e.g., polynomial or spline bases) and the parameter matrix $\Theta$ is unconstrained. By contrast, in NMF with covariates both $X$ and $\Theta$ are estimated under non-negativity constraints, 
which yields probabilistic and interpretable representations. Hence our setting can be viewed as a 
data-driven, interpretable extension of GCM. The factorization (\ref{XCA}) also belongs to the family of tri-NMF models 
\citep{ding2006}, with the important distinction that the covariate matrix $A$ 
is known in our setting. We call this the \textit{NMF with covariates model}, 
which corresponds to the forward problem. Importantly, this forward formulation 
has a natural dual, introduced in Section~\ref{sec3}, where the label matrix 
itself is treated as the observation. This duality highlights the conceptual 
connection between regression (forward) and classification (inverse) within 
the tri-NMF framework. A key feature of treating $A$ as known is that observation vectors are explained by individual-specific covariates, thereby allowing prediction for unseen covariate values. This property underlies applications of NMF with covariates using Gaussian-kernel covariates \citep{satoh2023}, 
growth curve modeling \citep{satoh2024}, and time-series VAR extensions \citep{satoh2025}.

We illustrate the forward problem using the orthodontic longitudinal dataset of \citet{potthoff1964}, 
available as \texttt{Orthodont} in the \texttt{nlme} package for R.  
The dataset consists of 27 children (16 boys and 11 girls) measured at ages 8\textendash14 years. Every two years, the distance between the pituitary and pterygomaxillary fissures was recorded. Here the observation matrix $Y$ contains the recorded \texttt{distance} values. With $P=4$ time points (ages 8, 10, 12, 14) and $N=27$ individuals, 
$Y$ is a $4 \times 27$ matrix whose $(p,n)$ entry is the distance for subject $n$ at age $p$, 
while the covariate matrix $A$ encodes sex as one-hot vectors, yielding a $2 \times 27$ matrix. Row and column labels are shown in (\ref{YandA}) for visual understanding, to clarify that 
each column of $Y$ and $A$ corresponds to the same individual and that male and female are 
represented by complementary $0\textendash1$ codings. \begin{align}
\!Y\! = \!
\begin{pmatrix}
    & \texttt{M01} & \scdots & \texttt{M16} & \texttt{F01} & \scdots & \texttt{F11} \\
8  & 26.0 & \scdots & 22.0 & 21.0 & \scdots & 24.5 \\
10 & 25.0 & \scdots & 21.5 & 20.0 & \scdots & 25.0 \\
12 & 29.0 & \scdots & 23.5 & 21.5 & \scdots & 28.0 \\
14 & 31.0 & \scdots & 25.0 & 23.0 & \scdots & 28.0 \\
\end{pmatrix},
A\!=\!
\begin{pmatrix}
       & \texttt{M01} & \scdots & \texttt{M16} & \texttt{F01} & \scdots & \texttt{F11} \\
\texttt{Male}   & 1 & \scdots & 1 & 0 & \scdots & 0 \\[3pt]
\texttt{Female} & 0 & \scdots & 0 & 1 & \scdots & 1 \\
\end{pmatrix}. \label{YandA}
\end{align}

\begin{figure}[h]
\centering
\includegraphics[width=0.85\linewidth]{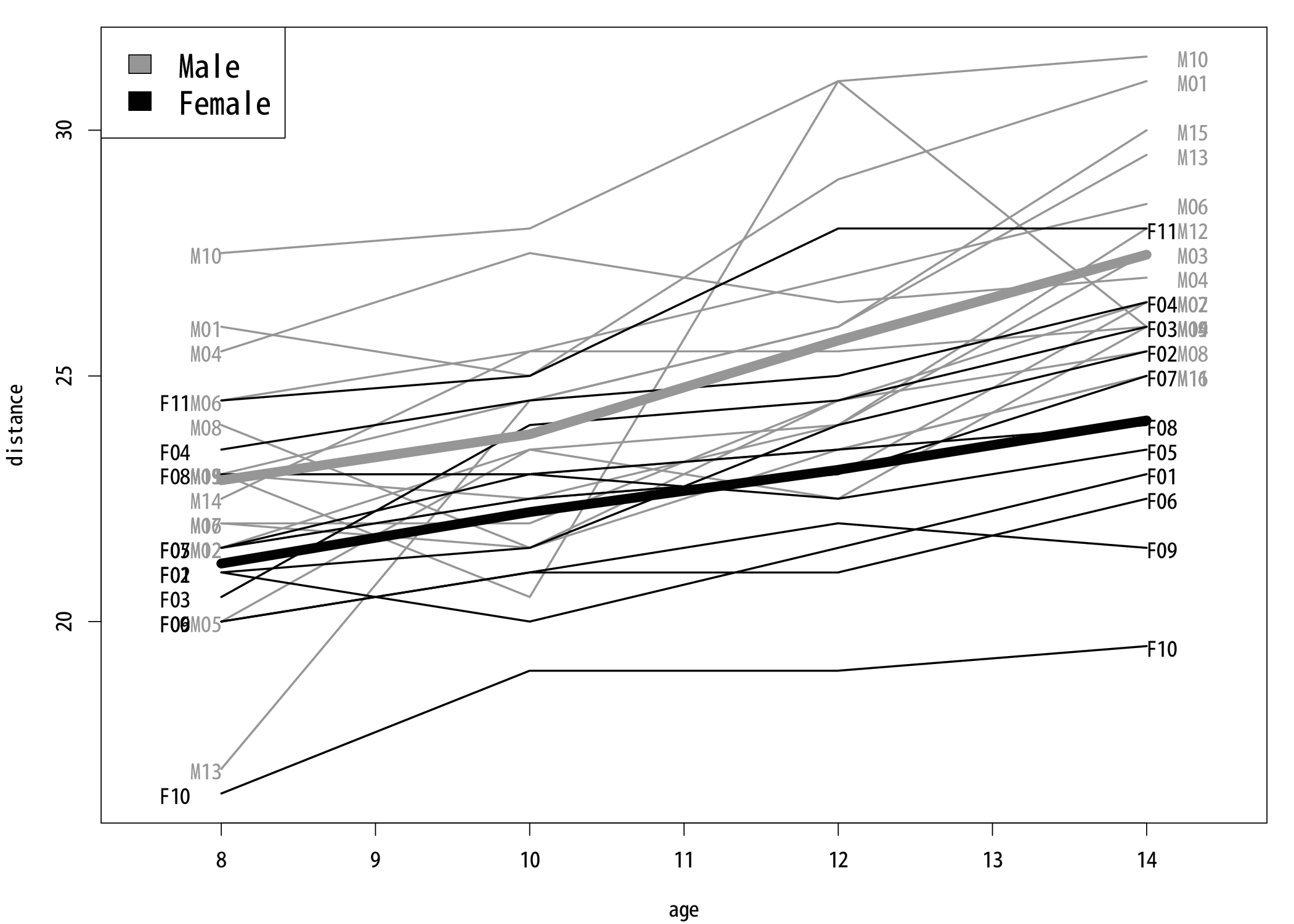}
\caption{Orthodontic longitudinal data analyzed by NMF with covariates. 
Thin lines denote individual trajectories, labeled with subject IDs at both ends. Bold curves indicate fitted trajectories for males (gray) and females (black)}
\label{fig1}
\end{figure}

Thus $Y$ contains longitudinal responses, while $A$ provides sex indicators. To compute fitted curves, we set $\bm{a}_n=(1,0)^\top$ for boys and $\bm{a}_n=(0,1)^\top$ for girls,  
where the symbol ${}^\top$ denotes vector or matrix transpose. Using the \texttt{nmfkc} package\footnote{\url{https://github.com/ksatohds/nmfkc}} 
for \textsf{R} \citep{satoh2024}, the estimated mean \texttt{distance} values were 
22.88, 23.81, 25.72, and 27.47 for boys, and 21.18, 22.23, 23.09, and 24.09 for 
girls, at ages 8, 10, 12, and 14, respectively. These fitted values capture 
sex-specific growth trajectories, as shown in Fig.~\ref{fig1}, where thin lines 
represent individual trajectories (subject IDs labeled at both ends) and bold 
curves depict the fitted trajectories for each sex. 

This example illustrates the forward problem. In Section~\ref{sec3}, we turn to the inverse formulation for classification. 
\section{NMF-LAB: Inverse problem of NMF with covariates for classification}\label{sec3}

In this section, we present the NMF-LAB model as the inverse formulation of NMF with covariates. Together with the forward problem in Section~\ref{sec2}, this model establishes a unified 
forward\textendash inverse duality within the tri-NMF framework. We first describe the representation of class labels using one-hot encoding and its probabilistic 
interpretation (Section~3.1). We then formulate the inverse problem within the tri-factorization 
framework (Section~3.2), and finally discuss normalization and identifiability issues that enable 
a probabilistic interpretation of the coefficients (Section~3.3).

\subsection{One-hot encoding and probabilistic interpretation}\label{sec31}

We represent class labels by one-hot vectors and discuss their probabilistic interpretation. Unlike Section~\ref{sec2}, where labels acted as covariates, 
they now form the observation matrix for classification. For sample $n$, the label vector is defined as 
$\bm{y}_n = (y_{1,n}, y_{2,n}, \dots, y_{P,n})^\top \in \{0,1\}^P$. It satisfies
\begin{equation}
y_{p,n} =
\begin{cases}
1 & \text{if sample $n$ belongs to class $p$,} \\[2pt]
0 & \text{otherwise},
\end{cases}
\qquad \text{and} \qquad
\sum_{p=1}^P y_{p,n} = 1.
\label{ylab}
\end{equation}

This corresponds to the standard one-hot encoding of class membership. Equivalently, $\bm{y}_n$ lies in the $P$-dimensional probability simplex
\begin{equation}
\Delta_P = \left\{ \bm{z} \in \mathbb{R}^P \;\middle|\; z_p \geq 0,\; \sum_{p=1}^P z_p = 1 \right\},
\label{prob}
\end{equation}
and represents a degenerate probability distribution concentrated on a single class. In the next subsection, we relax this representation to allow soft class memberships, 
which will be estimated from covariates via the NMF-LAB model.

\subsection{Tri-factorization formulation}\label{sec32}

In the NMF-LAB model, the observation matrix $Y=(\bm{y}_1,\dots,\bm{y}_N)$ 
is approximated by the tri-factorization model in Equation~(\ref{XCA}), namely 
$Y \approx X \Theta A$. We define the coefficient matrix as
\begin{equation}
B = \Theta A = (\bm{b}_1,\dots,\bm{b}_N) = (b_{q,n})_{Q \times N},
\label{Bdef}
\end{equation}
where each column $\bm{b}_n$ represents the coefficient vector for sample $n$. Then each $\bm{y}_n$ can be expressed as a linear combination of the column vectors of $X$:  
\begin{equation}
\bm{y}_n \approx b_{1,n} \bm{x}_1+\cdots+b_{Q,n} \bm{x}_Q,
\label{bx}
\end{equation}
where $\bm{x}_q$ denotes the $q$th column of the basis matrix $X$. If $Y$ is a label matrix (i.e., one-hot encoded class indicators), 
the basis matrix $X$ tends to have columns close to one-hot vectors 
once the columns are normalized to sum to one, 
because NMF inherently favors sparse representations. In this case, the coefficient vector $\bm{b}_n$ can naturally be interpreted as the class-membership probability vector, 
so the NMF-LAB formulation provides direct class probability estimates without an external classifier (e.g., logistic regression or SVM).

To reduce scale ambiguity and improve interpretability, 
we constrain each column of the basis matrix $X$ to sum to one. This column-stochastic normalization eliminates the arbitrary rescaling between $X$ and $B$ that is inherent in matrix factorization 
and also stabilizes the optimization procedure, as also described in \citep{satoh2025}. In addition, combined with the natural sparsity of NMF, column normalization supports partial identifiability \citep{gillis2023}. When the number of bases equals the number of classes ($Q=P$), 
the columns of $X$ often become close to one-hot label vectors up to permutation, 
so that $X$ effectively represents the class structure. While $X$ captures the class structure, 
the coefficient matrix $B$ requires normalization to yield valid probability distributions.

\subsection{Normalization and identifiability}\label{sec33}

In the NMF-LAB model, the coefficient matrix $B = (b_{q,n})$ encodes the degree of class
membership for each sample. As shown by comparing Equations~\eqref{XCa} and \eqref{bx}, 
the two representations imply that $\bm{b}_n = \Theta \bm{a}_n$. Each coefficient vector can then be normalized to obtain a probability distribution:
\begin{equation}
\tilde{b}_{q,n} = \frac{b_{q,n}}{\sum_{q'=1}^Q b_{q',n}}, 
\quad n = 1, \ldots, N,
\label{bprob}
\end{equation}
where $\tilde{\bm{b}}_n = (\tilde{b}_{1n}, \ldots, \tilde{b}_{Qn})^\top \in \Delta_Q$ represents the normalized
membership probabilities. Collecting these vectors yields the normalized coefficient
matrix $\tilde{B} = (\tilde{\bm{b}}_1, \ldots, \tilde{\bm{b}}_N) \in \mathbb{R}^{Q \times N}_+$. The reconstructed matrix 
\begin{equation}
\tilde{Y} = X \tilde{B}
\label{yhat}
\end{equation}
provides an estimate of the class-membership probability matrix, 
where each column of $\tilde{Y}$ is the estimated class-probability vector for a sample. This formulation has two key implications:  
(i) each column $\tilde{\bm{y}}_n \in \Delta_P$, so $\tilde{Y}$ directly functions as a probabilistic classifier; (ii) since the covariate matrix $A$ explicitly enters the factorization, the model naturally generalizes 
to unseen covariates, enabling prediction of class-probability distributions for new inputs.

In the forward problem (Section~\ref{sec2}), features are explained by covariates. In contrast, the inverse formulation treats the label matrix as the observation, thereby directly addressing classification. Taken together, these perspectives highlight the duality of the proposed framework, 
which unifies regression-type and classification-type problems.

Finally, column normalization aligns our framework with recent results on \textit{partial identifiability} \citep{gillis2023}. This theory provides a strong motivation for why the basis matrix $X$ in our model tends to be interpretable. It suggests that under certain conditions, such as the sparsity encouraged by our one-hot encoded label matrix $Y$, at least a subset of the latent factors can be reliably identified up to permutation and scaling. While the original theory addresses the exact two-factor NMF, and a rigorous proof for our approximate, covariate-constrained tri-NMF model remains an interesting open question for future research, this theoretical connection serves as a powerful justification for expecting the columns of $X$ to align closely with class indicators. Thus, the inverse tri-NMF formulation is supported not only by its practical utility in classification but also by a strong theoretical rationale that reinforces its interpretability and soundness.

\section{Implementation of the NMF-LAB model on Growth Curve Data}\label{sec4}

In this section, we apply the proposed NMF-LAB model to the orthodontic growth curve data. Section~4.1 reports the initial implementation and its limitations. 
Section~4.2 describes implementation aspects of kernel-based covariates, 
including the choice of kernel parameters and cross-validation strategy. Section~4.3 presents the theoretical foundation of kernel-based covariates 
based on the representer theorem. Section~4.4 demonstrates the improvement achieved using kernel matrices, 
while Section~4.5 evaluates the classification results. Finally, Section~4.6 discusses the optimization procedure and summarizes the implementation results. 
\subsection{Initial implementation and limitations}\label{sec41}
We demonstrate the implementation of the NMF-LAB model using the orthodontic growth curve data introduced in Section~2. In this example, we interchange the roles of the observation matrix $Y$ and the covariate matrix $A$, and optimize the model in Equation~(\ref{XCA}) with the number of bases fixed at $Q=P=2$. The computation was performed using the \texttt{nmfkc} package in R. 
However, the overall classification accuracy was only $40.7\%$, 
which is even lower than the $50\%$ chance level for a binary classification task. As shown in Equation~(\ref{Bdef}), this arises because each coefficient vector is constrained to be the product of the covariate vector and the parameter matrix, which limits model flexibility. To overcome this limitation, \citet{satoh2023,satoh2024} proposed using kernel matrices as covariates, which generally provide a better fit to the data. We therefore consider kernel-based covariates in the next subsection.

\subsection{Kernel-based covariates: implementation aspects}\label{sec42}

To improve model fit, we introduce kernel-based covariates. Following \citet{satoh2023}, we use the Gaussian (RBF) kernel, which is effective at
capturing nonlinear structures. Given individual data vectors 
$\bm{u}_1,\ldots,\bm{u}_N$, the kernel is
\begin{equation}
k(\bm{u}_n,\bm{u})=\exp\!\bigl(-\beta\,\|\bm{u}_n-\bm{u}\|^2\bigr), 
\qquad n=1,\ldots,N,
\label{bK}
\end{equation}
with a single hyperparameter $\beta>0$. When $\beta$ is too large, the kernel matrix
$A$ approaches identity and may overfit; when $\beta$ is too small, entries become
nearly uniform and lose discriminative power. We therefore optimize $\beta$ by 
cross-validation (CV) using the Frobenius loss (squared Euclidean loss). Importantly, when performing CV with a kernel matrix, one must remove not only the 
columns of the held-out samples but also the corresponding rows, in order to 
prevent information leakage from training to validation.

In practice, a common rule for selecting $\beta$ is the \textit{median heuristic}, 
which sets the bandwidth to the median of pairwise distances among samples. This simple strategy is widely used for its robustness \citep{gretton2012}, and its
large-sample properties have been analyzed theoretically \citep{garreau2017}. Although more elaborate optimization criteria exist, the median heuristic remains
a practical, computationally efficient baseline in large-scale applications. In our experiments, we adopt CV to fine-tune $\beta$ around this baseline.

The covariate vector for individual $n$ is then
\begin{equation}
\bm{a}_n=\bigl\{k(\bm{u}_1,\bm{u}_n),\ldots,k(\bm{u}_N,\bm{u}_n)\bigr\}^\top. \label{a}
\end{equation}
For a new feature vector $\bm{u}^{*}$, we build 
$\bm{a}^{*}=\{k(\bm{u}_{1},\bm{u}^{*}),\ldots,k(\bm{u}_{N},\bm{u}^{*})\}^\top$,
compute $\bm{b}^{*}=\Theta\bm{a}^{*}$ (normalized if necessary as in 
Eq.~(\ref{bprob})), and obtain the predicted probability 
$\tilde{\bm{y}}^{*}=X\bm{b}^{*}$, consistent with the probabilistic interpretation
in Section~\ref{sec33}. With the Gaussian-kernel covariates (Section~\ref{sec42}), the classification accuracy substantially improved
(details of the confusion matrices for both the direct and kernel designs are reported in Section~\ref{sec45}). 
\subsection{Theoretical foundation}\label{sec43}

The use of kernel-based covariates is supported by the representer theorem 
\citep{kimeldorf1971, scholkopf2001}. It states that the solution of certain 
regularized problems in reproducing kernel Hilbert spaces can be expressed as a 
finite linear combination of kernel functions evaluated at the training points:
\begin{equation}
b(u_n) \;=\; \sum_{m=1}^N \theta_m\, k(u_m, u_n),
\quad n = 1,\ldots,N,
\label{eq:rep}
\end{equation}
with expansion coefficients $\theta_m$. Stacking over $n$ and writing 
$K = (k(u_m,u_n))_{m,n=1}^N$, we obtain
\begin{equation}
B \;=\; \Theta_K\, K,
\label{eq:kernelB}
\end{equation}
where $B = (b(u_1), \ldots, b(u_N)) \in \mathbb{R}_+^{Q \times N}$ and
$\Theta_K \in \mathbb{R}_+^{Q \times N}$ is learned under the kernel-covariate design 
(here the number of covariates is $R=N$). Hence, Gaussian kernels provide a flexible mechanism to capture nonlinear relations 
between covariates and coefficients and can be viewed as a concrete instance of the 
representer theorem. The approach is closely related to kernel ridge regression 
\citep{murphy2012} and kernelized extensions of NMF \citep{chen2022}. Beyond the theoretical guarantee, the kernel induces a \emph{local averaging} effect:
each sample is represented relative to its neighbors via similarity weights. This facilitates nonlinear decision boundaries while smoothing predictions over nearby 
samples, which contributes to the empirical robustness of kernelized NMF-LAB to label 
noise observed in Section~\ref{sec5}. 
\subsection{Improved fit with kernel matrix}\label{sec44}

We applied the Gaussian-kernel-based covariates described in Section~\ref{sec42} 
to the orthodontic growth curve data. The kernel parameter was set to $\beta = 0.0079$, 
optimized using five-fold cross-validation (CV). \begin{eqnarray}
X =
\begin{pmatrix}
       & \texttt{Class1} & \texttt{Class2} \\
\texttt{Male}   & 1 & 0 \\
\texttt{Female} & 0 & 1 \\
\end{pmatrix}
\label{X}
\end{eqnarray}

The estimated basis matrix $X$ coincides with a permuted identity matrix, 
which corresponds to the label matrix, assigning one basis to each sex category (male or female). For convenience, the two bases are labeled as \texttt{Class1} and \texttt{Class2}. Since $X$ in Equation~(\ref{X}) is the identity matrix and $XB$ can be interpreted as membership probabilities, 
the coefficient matrix $B$ can also be regarded as an approximate class-membership probability matrix. As described in Equation~(\ref{bx}), its column sums are expected to be close to one. In fact, the average column sum over the 27 individuals was $1.036$, with a standard deviation of $0.080$, 
indicating that the raw coefficients already approximate valid probability vectors. Next, using Equation~(\ref{bprob}), we adjusted each coefficient vector so that its components summed exactly to one. This yielded the normalized coefficient matrix $\tilde{B}$, whose columns represent valid probability distributions over the classes. The class-membership probability matrix was then obtained as
\begin{eqnarray}
\tilde{Y} = X\tilde{B} =
\begin{pmatrix}
       & \texttt{M01} & \scdots & \texttt{M16} & \texttt{F01} & \scdots & \texttt{F11} \\
\texttt{Male}   & 0.94 & \scdots & 0.49 & 0.28 & \scdots & 0.86 \\
\texttt{Female} & 0.06 & \scdots & 0.51 & 0.72 & \scdots & 0.14 \\
\end{pmatrix}. \label{Yhatprob}
\end{eqnarray}

\subsection{Classification results and evaluation}\label{sec45}

Based on the class-membership probabilities in Equation~(\ref{Yhatprob}), 
each subject was assigned to the class with the higher probability. Table~\ref{Tab44} summarizes the classification results for both the 
direct covariates (linear $B=\Theta A$) and the kernel covariates 
(Gaussian kernel, $\beta$ tuned by CV). Rows correspond to predicted labels and columns correspond to true labels.

\begin{table}[htbp]
\captionsetup{width=\textwidth}
\centering
\caption{Confusion matrices on the orthodontic dataset under the same evaluation protocol. Left: direct covariates (linear $B=\Theta A$, accuracy $40.7\%$). Right: kernel covariates (Gaussian kernel; $\beta$ tuned by CV, accuracy $77.8\%$)}
\label{Tab44}
\begin{tabular}{|c|cc|cc|}
\multicolumn{1}{c}{} &\multicolumn{2}{c}{direct covariates} & \multicolumn{2}{c}{kernel covariates} \\
\hline
 & \texttt{Female} & \texttt{Male}  & \texttt{Female} & \texttt{Male} \\
\hline
Pred.\ \texttt{Female} & 0  & 0   & 7  & 2 \\
Pred.\ \texttt{Male}   & 11 & 16  & 4  & 14 \\
\hline
\end{tabular}
\end{table}

\subsection{Optimization}\label{sec46}

For the optimization of NMF-LAB, we adopt multiplicative update rules derived within the tri-NMF framework, as discussed in \citet{satoh2025}. Given the observation matrix $Y$ (the label matrix), the covariate matrix $A$, and the approximation $\hat{Y}=X\Theta A$, the basis matrix $X$ and parameter matrix $\Theta$ are updated as follows:
\begin{eqnarray}
X \longleftarrow X \odot (YA^\top\Theta^\top \oslash \hat{Y}A^\top \Theta^\top), \qquad
\Theta  \longleftarrow \Theta \odot \{(X^\top Y A^\top) \oslash (X^\top \hat{Y} A^\top)\}, 
\label{renewtheta_lab}
\end{eqnarray}
where $\odot$ and $\oslash$ denote Hadamard product and division, respectively. After each update, we normalize the columns of $X$ to sum to one, which allows them to be interpreted as probability vectors representing class membership. These updates guarantee a monotonic decrease of the squared Euclidean loss $D_{EU}(Y,\hat{Y})$ following the auxiliary function approach of \citet{lee1999,lee2000}. As the NMF objective function is non-convex, they are guaranteed to converge only to a locally optimal solution, but do so with numerical stability \citep{ding2006}.

Initialization plays a crucial role since the convergence to a local minimum makes multiplicative updates sensitive to starting points \citep{gillis2023}. A common choice is to use K-means centroids of $Y$, which improves convergence speed and interpretability \citep{fathi2023}. In the present setting, however, $Y$ is (or is close to) a label matrix: when $Y$ is fully labeled, $X$ coincides with the identity matrix (up to permutation), and even in the semi-supervised case of Section~\ref{sec:robustness}, where $Y$ contains probability vectors for unlabeled samples, $X$ remains close to the identity matrix. This provides a particularly stable and interpretable initialization, aligning basis vectors with class labels from the start, accelerating convergence, and reducing the risk of poor local minima. In practice, this initialization strategy has been found to improve both accuracy and efficiency in classification tasks, and will be adopted in the experiments presented in Section~\ref{sec:empirical_evaluation}. In summary, the optimization of NMF-LAB ensures monotonic convergence, 
offers stable and interpretable initialization (K-means in general or the identity matrix in classification tasks), 
and enhances both accuracy and efficiency, yielding a robust and practically useful framework for classification.

\section{Empirical Evaluation and Performance Analysis}
\label{sec:empirical_evaluation}

In this chapter, we evaluate the effectiveness of the proposed Non-negative Matrix Factorization for Label Matrix (\textit{NMF-LAB}) through classification tasks using diverse real-world datasets. Specifically, we investigate whether NMF-LAB with kernel covariates (\textit{NMF-LAB (kernel)}) can achieve a favorable balance between the interpretability of a linear model (\textit{NMF-LAB (direct)}) and the high predictive performance and robustness to label noise characteristic of non-linear models. We first outline the overall experimental design, including the datasets, comparison methods, and evaluation protocols (Section~\ref{sec:exp_setup}). Next, we assess the overall classification performance on small- to medium-scale benchmark datasets (Section~\ref{sec:benchmark_perf}). We then conduct detailed analyses on robustness to label noise (Section~\ref{sec:robustness}) and the interpretability of the linear model (Section~\ref{sec:interpretability}) using specific case studies. Finally, we demonstrate the scalability of our approach using the large-scale MNIST dataset with the Nyström approximation (Section~\ref{sec:scalability}). 
\subsection{Experimental Setup}
\label{sec:exp_setup}

\subsubsection{Datasets}
Our evaluation employs eight real-world datasets with varying characteristics. Seven are small- to medium-scale benchmarks (RBGlass1, Iris, Penguins, Wine, Seeds, Vehicle, and Digits), while MNIST serves as a large-scale benchmark. A summary of their properties is provided in Table~\ref{tab:datasets_detail}. For all experiments, feature matrices (covariates $A$) were column-normalized to the range $[0, 1]$ to satisfy the non-negativity constraint, and samples with missing values were excluded. \begin{table}[ht]
\captionsetup{width=\textwidth}
\centering
\caption{Summary of Datasets Used in Numerical Experiments}
\label{tab:datasets_detail}
\begin{tabular}{l|c|c|c}
\hline
\textit{Dataset} & \textit{Samples ($N$)} & \textit{Features ($R$)} & \textit{Classes ($P$)} \\
\hline
RBGlass1 & 105 & 11 & 2 \\
Iris & 150 & 4 & 3 \\
Penguins & 333 & 4 & 3 \\
Wine & 178 & 13 & 3 \\
Seeds & 199 & 7 & 3 \\
Vehicle & 846 & 18 & 4 \\
Digits & 1,797 & 64 & 10 \\
MNIST (Train/Test) & 60,000 / 10,000 & 784 & 10 \\
\hline
\end{tabular}
\end{table}

\subsubsection{Data Splitting Protocol}
\begin{itemize}
    \item \textit{Small- to Medium-Scale Datasets:} For the seven datasets from RBGlass1 to Digits, we used stratified sampling to split 
the data into training (40\%), validation (40\%), and test (20\%) sets, preserving class proportions. This procedure was repeated 50 times with different random seeds to compute the mean and standard deviation of the test accuracy.
    \item \textit{Large-Scale Dataset (MNIST):} For MNIST, we randomly selected 10,000 images from the 60,000 training samples for hyperparameter tuning and 2,000 for validation. The final model was trained on the full 60,000 training images and evaluated on the dedicated 10,000 test images. This experiment was repeated 10 times.
\end{itemize}

\subsubsection{Hyperparameter Optimization}
To ensure a fair comparison, all methods underwent hyperparameter optimization (HPO). HPO was performed using the training and validation sets, selecting the parameters that yielded the highest accuracy or lowest loss on the validation set. The final model was then retrained on the combined training and validation data before being evaluated on the test set. \begin{itemize}
    \item \textit{NMF-LAB (direct and kernel):} 
    The latent dimension $Q$ was set equal to the number of classes $P$. The objective function for NMF-LAB is to find $X$ and $\Theta$ that minimize the reconstruction error of the label matrix $Y$ from the covariate matrix $A$:
    \begin{equation}
        \min_{X, \Theta} \| Y - X \Theta A \|^2.
    \end{equation}
    For the \textit{NMF-LAB (kernel)} variant, the Gaussian kernel parameter $\beta$ was optimized over four candidates, $\beta_{\text{median}} \times \{10^{-2}, 10^{-1}, 10^{0}, 10^{1}\}$, centered around the value derived from the median heuristic.
    \item \textit{SSNMF:}
    The SSNMF model used for comparison follows a standard joint factorization approach, which is fundamentally different from the inverse problem formulation of NMF-LAB. It seeks to simultaneously reconstruct the feature matrix $A$ and the label matrix $Y$ using a shared latent representation matrix $S$. The objective function is:
    \begin{equation}
        \min_{X_A, X_Y, S} \| A - X_A S \|^2 + \lambda \| Y - X_Y S \|^2,
    \end{equation}
    where $X_A$ and $X_Y$ are the basis matrices for features and labels, respectively. The trade-off parameter $\lambda$, which balances the reconstruction of features and labels, was optimized via grid search over $\{0.1, 1, 10\}$.
    \item \textit{Baseline Classifiers:} Parameters for NN, SVM, MLR, RF, KNN, and CART were tuned using the \texttt{caret} package in R, with search grids as specified below.
    \begin{itemize}
        \item \textit{Neural Network (NN):} A single-hidden-layer network was used, where the number of hidden units (\texttt{size}) was set equal to the number of classes $P$. The weight decay parameter (\texttt{decay}) was optimized over $\{10^{-3}, 10^{-2}, 10^{-1}\}$. Training was performed in chunks (\texttt{maxit=100} per chunk, up to 6 chunks) with early stopping.
        \item \textit{Support Vector Machine (SVM):} With a Radial Basis Function (RBF) kernel, the cost parameter $C \in \{0.1, 1, 10\}$ and kernel width $\sigma \in \{0.01, 0.1, 1\}$ were tuned.
        \item \textit{Multinomial Logistic Regression (MLR):} Using a Ridge penalty ($\alpha=0$), the regularization strength $\lambda$ was tuned over 60 logarithmic steps between $1$ and $10^{-4}$.
        \item \textit{Random Forest (RF):} The number of variables randomly sampled at each split (\texttt{mtry}) was tuned over $\{1, 2, 3, 4\}$.
        \item \textit{k-Nearest Neighbors (KNN):} The number of neighbors ($k$) was tuned over the range $1$ to $10$.
        \item \textit{CART:} The complexity parameter (\texttt{cp}) was tuned over $\{0.001, 0.01, 0.05, 0.1\}$.
    \end{itemize}
\end{itemize}

\subsection{Overall Classification Performance on Standard Benchmarks}
\label{sec:benchmark_perf}

We first compare the performance of all methods on the seven small- to medium-scale datasets. Table~\ref{Tab:hard_comparison} summarizes the test accuracy (mean $\pm$ standard deviation) under the hard label setting (i.e., all labels are perfectly correct).
\begin{table}[ht]
  \centering
  \small 
  \setlength{\tabcolsep}{4pt} 
  \caption{Comparison of Test Accuracy for Classifiers on Each Dataset (\%, Mean $\pm$ Standard Deviation). `Direct` and `Kernel` refer to NMF-LAB (direct) and NMF-LAB (kernel), respectively}
  \label{Tab:hard_comparison}
  \begin{tabular}{l|rrrrrrr}
    \hline
    & Glass & Iris & Penguins & Wine & Seeds & Vehicle & Digits \\
    \hline
    Direct & $83.0 \pm 7.8$ & $66.7 \pm 0.0$ & $79.1 \pm 0.2$ & $88.0 \pm 5.1$ & $82.7 \pm 4.9$ & $52.5 \pm 3.9$ & $68.0 \pm 2.6$ \\
    Kernel & $84.3 \pm 9.3$ & $95.5 \pm 3.6$ & $97.9 \pm 1.5$ & $96.2 \pm 2.9$ & $94.2 \pm 3.2$ & $69.3 \pm 2.8$ & $94.6 \pm 1.1$ \\
    SSNMF  & $83.4 \pm 7.7$ & $69.1 \pm 5.1$ & $99.3 \pm 1.0$ & $91.2 \pm 5.0$ & $88.2 \pm 4.7$ & $43.2 \pm 5.2$ & $88.4 \pm 1.4$ \\
    NN     & $86.0 \pm 6.5$ & $96.7 \pm 3.1$ & $98.8 \pm 1.2$ & $97.6 \pm 2.1$ & $95.9 \pm 3.0$ & $81.7 \pm 2.4$ & $96.8 \pm 1.0$ \\
    MLR    & $84.6 \pm 7.3$ & $94.1 \pm 4.2$ & $97.3 \pm 1.9$ & $97.2 \pm 2.5$ & $93.1 \pm 3.6$ & $74.7 \pm 2.4$ & $95.9 \pm 0.9$ \\
    SVM    & $78.2 \pm 15.0$ & $95.5 \pm 3.9$ & $97.4 \pm 2.1$ & $97.6 \pm 2.5$ & $93.6 \pm 3.6$ & $80.6 \pm 2.4$ & $96.2 \pm 2.1$ \\
    CART   & $80.2 \pm 7.2$ & $94.2 \pm 3.4$ & $93.1 \pm 2.4$ & $83.1 \pm 5.7$ & $90.4 \pm 3.8$ & $67.8 \pm 3.4$ & $82.2 \pm 2.6$ \\
    RF     & $88.9 \pm 6.6$ & $95.1 \pm 3.4$ & $97.9 \pm 1.4$ & $98.2 \pm 1.9$ & $93.6 \pm 3.6$ & $74.7 \pm 2.6$ & $97.7 \pm 0.8$ \\
    KNN    & $85.2 \pm 7.0$ & $95.6 \pm 3.3$ & $98.4 \pm 1.3$ & $95.0 \pm 2.7$ & $93.4 \pm 3.5$ & $71.1 \pm 2.3$ & $97.5 \pm 0.7$ \\
    \hline
  \end{tabular}
\end{table}

\textit{Key Insights.}
The most important empirical finding from these results is the stark performance gap between the linear NMF-LAB (direct) and the non-linear NMF-LAB (kernel) across most datasets. This gap highlights that the effectiveness of the direct model is highly dependent on the linear separability of the data. For instance, the 66.7\% accuracy of the direct model on the Iris dataset strongly suggests a failure to distinguish between the two classes that are not linearly separable, a task for which a linear model is fundamentally unsuited.

This result underscores a central finding of our work: while the direct model offers interpretability (as explored in Section~\ref{sec:interpretability}), its predictive power is limited to linearly separable problems. In contrast, the NMF-LAB (kernel) model demonstrates that by incorporating a non-linear mapping via the kernel, the framework achieves high performance comparable to powerful baselines like Neural Networks and Random Forests, even on complex datasets. This confirms that the non-linear kernel extension is essential for the NMF-LAB framework to be competitive in general classification tasks and clearly illustrates the fundamental trade-off between the direct model's interpretability and the kernel model's predictive accuracy.

\subsection{Robustness to Label Noise}
\label{sec:robustness}

To evaluate the robustness of the NMF-LAB framework against noisy supervision, we conducted an experiment using soft labels. The soft labeling protocol introduces controlled label uncertainty only into the training data. For a sample belonging to the true class $c^\ast$, a probability $r$ is assigned to $c^\ast$, and the remaining probability $(1-r)$ is uniformly distributed among the other $(P-1)$ classes. We used the Iris and Seeds datasets for this evaluation. Tables~\ref{Tab:robust_iris} and \ref{Tab:robust_seeds} report the test accuracy under varying soft label ratios.
\begin{table}[ht]
\captionsetup{width=\textwidth}
  \centering
  \caption{Test Accuracy (\%) on Iris Dataset under Soft Label Noise. The first column represents $r \times 100 (\%)$}
  \label{Tab:robust_iris}
  \begin{tabular}{l|rrrr}
    \hline
    $r \times 100 (\%)$ & NMF-LAB (direct) & NMF-LAB (kernel) & SSNMF & NN \\ 
    \hline
    0 & $66.7 \pm 2.9$ & $82.3 \pm 6.0$ & $71.5 \pm 5.8$ & $87.5 \pm 8.0$ \\ 
    20 & $66.4 \pm 1.3$ & $85.0 \pm 5.1$ & $75.0 \pm 6.6$ & $94.3 \pm 3.5$ \\ 
    40 & $66.5 \pm 0.9$ & $95.8 \pm 4.0$ & $72.8 \pm 6.0$ & $93.9 \pm 4.8$ \\ 
    60 & $66.7 \pm 0.7$ & $96.1 \pm 3.7$ & $70.9 \pm 5.4$ & $96.6 \pm 3.1$ \\ 
    80 & $66.7 \pm 0.0$ & $95.8 \pm 3.5$ & $70.9 \pm 5.7$ & $96.9 \pm 2.8$ \\ 
    100 & $66.7 \pm 0.0$ & $95.5 \pm 3.6$ & $69.1 \pm 5.1$ & $96.7 \pm 3.1$ \\ 
    \hline
  \end{tabular}
\end{table}

\begin{table}[ht]
\captionsetup{width=\textwidth}
  \centering
  \caption{Test Accuracy (\%) on Seeds Dataset under Soft Label Noise. The first column represents $r \times 100 (\%)$}
  \label{Tab:robust_seeds}
  \begin{tabular}{l|rrrr}
    \hline
    $r \times 100 (\%)$ & NMF-LAB (direct) & NMF-LAB (kernel) & SSNMF & NN \\ 
    \hline
    0 & $83.9 \pm 6.6$ & $81.4 \pm 4.8$ & $63.1 \pm 11.4$ & $83.7 \pm 7.7$ \\ 
    20 & $86.1 \pm 5.3$ & $86.4 \pm 4.3$ & $79.7 \pm 10.8$ & $91.7 \pm 4.4$ \\ 
    40 & $85.8 \pm 5.0$ & $93.7 \pm 3.6$ & $83.1 \pm 7.3$ & $92.3 \pm 3.7$ \\ 
    60 & $85.2 \pm 4.6$ & $94.1 \pm 3.5$ & $87.1 \pm 4.7$ & $94.9 \pm 3.6$ \\ 
    80 & $84.6 \pm 4.6$ & $94.6 \pm 3.3$ & $87.7 \pm 4.7$ & $95.2 \pm 3.6$ \\ 
    100 & $82.7 \pm 4.9$ & $94.2 \pm 3.2$ & $88.2 \pm 4.7$ & $95.9 \pm 3.0$ \\ 
    \hline
  \end{tabular}
\end{table}

\textit{Key Insights.}
The results largely support the claim that the NMF-LAB framework is robust to label noise, likely due to the inherent smoothing effect of the factorization process. This is particularly evident in the Seeds dataset (Table~\ref{Tab:robust_seeds}), where the accuracy of NMF-LAB (kernel) consistently improves as the label quality increases (i.e., as $r$ increases), significantly outperforming the direct and SSNMF models.

However, the results on the Iris dataset (Table~\ref{Tab:robust_iris}) reveal a more nuanced picture. While NMF-LAB (kernel) performs well under moderate noise levels, it is surpassed by the Neural Network in the extreme case of completely random labels ($r=0$). A plausible explanation is that the kernel's local averaging mechanism, while effective at smoothing out partial noise, may be misled by uniformly random label information. In contrast, a neural network might have a greater capacity to learn more abstract representations that are invariant to such unstructured noise. This highlights that while NMF-LAB demonstrates considerable robustness, its performance relative to other non-linear models can depend on the specific nature and severity of the label noise.

\subsection{Interpretability Case Study: The RBGlass1 Dataset}
\label{sec:interpretability}

While NMF-LAB (kernel) generally provides superior predictive accuracy, the linear version, NMF-LAB (direct), offers significant advantages in model interpretability. We select the RBGlass1 dataset for this case study because, as shown in Table~\ref{Tab:hard_comparison}, it was one of the few datasets where its performance was comparable to that of other baseline methods. For this analysis, which focuses on interpretability rather than predictive generalization, the parameter matrix $\Theta$ was estimated using all available data for the RBGlass1 dataset to obtain the most stable model for interpretation.

\begin{figure}[h]
\centering
\includegraphics[width=0.85\linewidth]{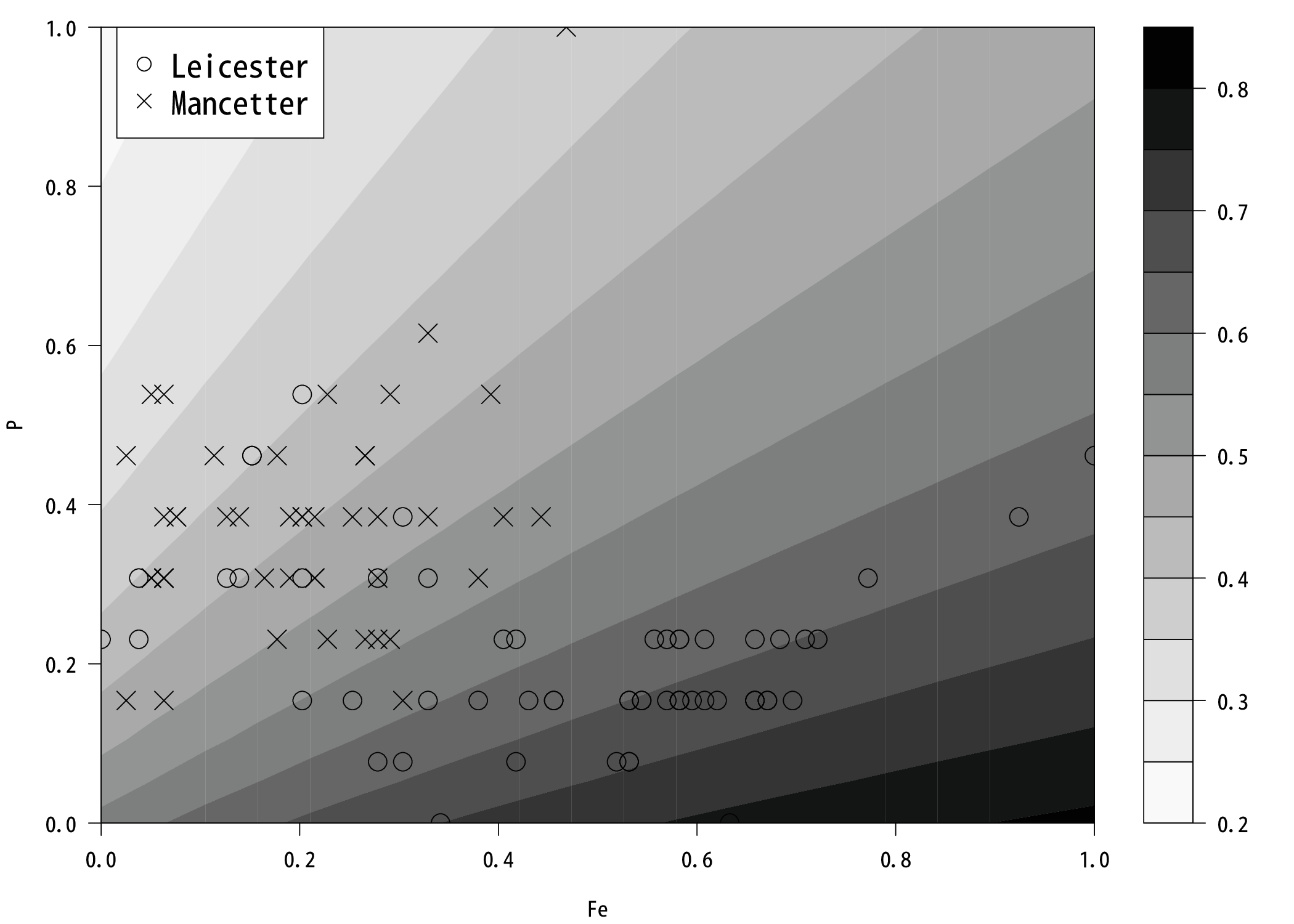}
\caption{Predicted probability of \texttt{Leicester} as a function of \texttt{Fe} (horizontal axis) and \texttt{P} (vertical axis), with all other covariates fixed at their mean values. Darker shading indicates higher probability. Circles denote samples from \texttt{Leicester}, and crosses denote samples from \texttt{Mancetter}}
\label{fig:FeP_Interpretation}
\end{figure}

The direct formulation, $B = \Theta A$, uses $\Theta$ to reveal the direct, non-negative, and additive contributions of original features ($A$) to the class membership probabilities. The estimated parameter matrix is shown in Equation~\ref{Eq:Theta}, and the predicted probability of the `Leicester` class is visualized in Fig.~\ref{fig:FeP_Interpretation}. The visualization clearly demonstrates the model's interpretability; for example, the probability of belonging to the `Leicester` class increases with the value of Fe, while the probability of the `Mancetter` class increases with P and Mn. This clear, parts-based interpretation is a classic strength of NMF.

\begin{eqnarray}
\captionsetup{width=\textwidth}
\Theta \!= \!\left(
\begin{array}{cccccccccccc}
\, & \texttt{Al} & \texttt{Fe} & \texttt{Mg} & \texttt{Ca} & \texttt{Na} & \texttt{K} & \texttt{Ti} & \texttt{P} & \texttt{Mn} & \texttt{Sb} & \texttt{Pb}\\
\texttt{Leicester}  & 0     & 0.73 & 0 & 0.16 & 0 & 0 & 0.07 & 0 & 0 & 0.54 & 0 \\ 
\texttt{Mancetter} & 0.02 & 0     & 0 & 0.17 & 0 & 0 & 0 & 0.88 & 0.56 & 0 & 0 \\ 
\end{array}
\right). \label{Eq:Theta}
\end{eqnarray}

However, it is crucial to recognize that this compelling interpretability is derived from a model that, as demonstrated in Section~\ref{sec:benchmark_perf}, fails to achieve competitive accuracy on most datasets. This raises a fundamental question regarding the practical value of interpretability when predictive performance is lacking. This case study, therefore, serves to powerfully illustrate the central trade-off that this paper uncovers: to achieve the high predictive accuracy of the kernel model, one must sacrifice the direct, feature-level interpretability that the linear direct model provides. This trade-off is not a limitation but a core finding of our investigation into the NMF-LAB framework, highlighting the explicit choice between a transparent but simple model and a powerful but opaque one.

\subsection{Scalability to Large-Scale Data: MNIST}
\label{sec:scalability}

The final experiment addresses the scalability of NMF-LAB, a critical consideration for its practical application. We evaluate the kernel-based approach on the large-scale MNIST handwritten digit dataset, where the computational cost of the full kernel matrix is prohibitive. This section first details the Nyström approximation, the key technique used to make the kernel method tractable for such large datasets (Section~\ref{sec:nystrom_details}). We then present the classification results, which demonstrate the model's ability to scale effectively while maintaining competitive accuracy (Section~\ref{sec:performance_mnist}). 
\subsubsection{Nyström Approximation for Kernel Matrix}
\label{sec:nystrom_details}

Given the large number of training samples ($N=60,000$), direct computation of the full $N \times N$ Gaussian kernel matrix is computationally infeasible. To address this issue, we employ the Nyström method \citep{nystrom1930,williams2000} to approximate the kernel matrix. Let $\bm{u}_1,\ldots,\bm{u}_N$ denote the observed feature vectors. We first select $M \ll N$ landmark points, $\bm{v}_1,\ldots,\bm{v}_M$, which are obtained as the centroids of $k$-means clustering on a subset of the data. The full kernel matrix $K$ is defined as:
\begin{equation}
\mathop{K}_{N\times N} = 
\begin{pmatrix}
k(\bm{u}_1,\bm{u}_1) & \cdots & k(\bm{u}_1,\bm{u}_N)\\
\vdots & \ddots & \vdots \\
k(\bm{u}_N,\bm{u}_1) & \cdots & k(\bm{u}_N,\bm{u}_N)
\end{pmatrix}. \end{equation}
The Nyström method approximates $K$ as $K \approx C W^{-1} C^\top$, where the matrices $C$ and $W$ are defined as:
\begin{align}
\mathop{C}_{N\times M} &=
\begin{pmatrix}
k(\bm{u}_1,\bm{v}_1) & \cdots & k(\bm{u}_1,\bm{v}_M)\\
\vdots & \ddots & \vdots \\
k(\bm{u}_N,\bm{v}_1) & \cdots & k(\bm{u}_N,\bm{v}_M)
\end{pmatrix}, \\
\mathop{W}_{M\times M} &=
\begin{pmatrix}
k(\bm{v}_1,\bm{v}_1) & \cdots & k(\bm{v}_1,\bm{v}_M)\\
\vdots & \ddots & \vdots \\
k(\bm{v}_M,\bm{v}_1) & \cdots & k(\bm{v}_M,\bm{v}_M)
\end{pmatrix}. \end{align}
Here, $C \in \mathbb{R}^{N \times M}$ contains the kernel evaluations between all data points and the landmarks, while $W \in \mathbb{R}^{M \times M}$ is the kernel matrix of the landmarks themselves. According to \citet{zhang2008}, the approximation error is closely related to the quantization error of the landmark selection, and choosing landmarks as the $k$-means centroids provides both theoretical guarantees and practical accuracy. Following the formulation in Section~\ref{sec43}, where $B = \Theta_K K$, we use the Nyström approximation to write $B \approx \Theta_K (C W^{-1} C^\top)$. By defining a new parameter matrix $\Theta_C = \Theta_K C W^{-1}$, this simplifies to $B \approx \Theta_C C^\top$. Thus, in the MNIST experiment, we replace the full covariate matrix $A$ with the reduced covariate matrix $C^\top$ and optimize the corresponding parameter matrix $\Theta_C$. This approach reduces the computational complexity from $O(N^2)$ to $O(NM)$.

\subsubsection{Performance Evaluation on MNIST}
\label{sec:performance_mnist}

Table~\ref{Tab:hard_mnist_comparison} summarizes the hard-label classification accuracy on the MNIST test set. \textit{NMF-LAB (kernel500)} and \textit{NMF-LAB (kernel1000)} correspond to Nyström approximations with 500 and 1000 landmarks, respectively. \textit{NMF-LAB (full kernel)} refers to the original kernel implementation without Nyström approximation. \begin{table}[ht]
\captionsetup{width=\textwidth}
  \centering
  \caption{Test Accuracy (\%) on MNIST Dataset (Hard Labels, Mean $\pm$ SD over 10 runs)}
  \label{Tab:hard_mnist_comparison}
  \begin{tabular}{l|c}
    \hline
    \textit{Method} & \textit{Accuracy (\%)} \\
    \hline
    \textit{Proposed Methods} & \\
    NMF-LAB (direct) & $69.1 \pm 0.0$ \\
    NMF-LAB (kernel500) & $91.1 \pm 0.0$ \\
    NMF-LAB (kernel1000) & $93.7 \pm 0.0$ \\
    NMF-LAB (full kernel) & $96.0 \pm 0.0$ \\
    \hline
    \textit{Baseline Methods} & \\
    
    SSNMF & $66.4 \pm 0.0$ \\
    NN & $88.0 \pm 0.7$ \\
    MLR & $92.1 \pm 0.1$ \\
    SVM & $86.4 \pm 0.4$ \\
    CART & $79.9 \pm 0.0$ \\
    RF & $95.8 \pm 0.0$ \\
    KNN & $94.4 \pm 0.1$ \\
    \hline
  \end{tabular}
\end{table}

\textit{Key Insights.}
The kernelized NMF-LAB models vastly outperformed the linear versions (NMF-LAB (direct) at 69.1\% and SSNMF at 66.4\%), confirming that kernelization is essential for complex datasets like MNIST. Accuracy consistently improves with the number of landmarks, from 91.1\% (500 landmarks) to 93.7\% (1000 landmarks) and finally 96.0\% (full kernel), demonstrating a clear relationship between approximation quality and performance. The full kernel NMF-LAB (96.0\%) performs comparably to the powerful Random Forest baseline (95.8\%) and surpasses KNN (94.4\%). This result validates that the NMF-LAB framework, via kernel approximation, can scale efficiently to large, high-dimensional data while maintaining competitive accuracy.

\section{Conclusion and Discussion}\label{sec6}

In this study, we proposed \textit{NMF-LAB}, which formulates classification as the inverse problem of non-negative matrix tri-factorization (tri-NMF; \citealp{ding2006}). In this framework, the label matrix $Y$ is directly factorized with covariates $A$, enabling class probabilities to be obtained without an external classifier. Column normalization makes the basis matrix $X$ align with class indicators, while the coefficients represent class membership probabilities. By incorporating covariates such as Gaussian kernel functions \citep{satoh2023,satoh2024}, NMF-LAB can flexibly model nonlinear structures, generalize to unseen data, and naturally handle unlabeled samples, making it suitable for both supervised and semi-supervised learning.

Experiments on diverse datasets demonstrated that NMF-LAB achieves a favorable balance between predictive accuracy, interpretability, and flexibility. Importantly, the method exhibited robustness to label noise, and its effectiveness was validated on datasets ranging from small-scale benchmarks like Iris to the large-scale MNIST handwritten digit dataset, where its scalability was successfully demonstrated using the Nyström approximation.

The novelty of this study lies in treating the label matrix itself as the direct target of factorization and in casting classification as the inverse problem of tri-NMF. Existing supervised and semi-supervised NMF approaches \citep{leuschner2018,wang2015,wu2015} incorporate labels only as constraints or penalties. In contrast, multi-label learning methods \citep{yu2014,zhang2015} primarily exploit correlations among labels to improve prediction accuracy or impute missing labels, focusing on the internal structure of the label space rather than on mapping new covariates to label probabilities. By contrast, the proposed NMF-LAB framework reconstructs the label matrix $Y$ directly from the covariate matrix $A$, and therefore enables immediate estimation of class membership probabilities for new observations. Moreover, while earlier studies \citep{satoh2023,satoh2024,satoh2025} addressed the forward problem of NMF with covariates, the present work introduces the inverse setting, thereby establishing a duality between forward regression-type problems and inverse classification problems within the tri-NMF framework.

Several limitations remain. A primary point is the trade-off between performance and transparency; kernel-based covariates enhance predictive accuracy but reduce the direct feature-level interpretability of the linear model. While our proposed initialization strategy for classification tasks is effective, the NMF objective function is non-convex, meaning the multiplicative updates only guarantee convergence to a local optimum. Furthermore, our entire framework is built upon the squared Euclidean loss; exploring other loss functions, such as the Kullback-Leibler divergence, could be a valuable direction for future work. A promising approach to mitigate the interpretability trade-off could involve using deep neural networks to extract meaningful intermediate representations for use as covariates $A$ in the tri-factorization $Y \approx X \Theta A$.

Another challenge is scalability. While this study successfully demonstrated scalability to a large number of samples ($N=60,000$) via the Nyström approximation, its application to datasets with extremely high-dimensional features may still pose computational challenges. Future research could explore other approximation techniques, such as random feature methods \citep{rahimi2007}, to further enhance efficiency.

In addition, although this study focused on multi-class classification, extensions to multi-label learning and to settings with missing labels remain important directions. In particular, comparisons with frameworks for large-scale multi-label learning, such as \citet{yu2014}, will be essential. Moreover, in this work we set the number of bases $Q$ equal to the number of classes $P$. Developing methods that achieve good classification performance with fewer bases than classes ($Q < P$), for example to encourage compact representations, remains an important avenue for future research.

In summary, NMF-LAB provides a novel framing of classification as the inverse problem of tri-NMF, offering a unified, probabilistic, and interpretable framework for classification. By highlighting the duality between forward and inverse problems, the study contributes both a theoretical advance and a practical foundation for extending NMF into supervised learning contexts.

Looking ahead, a promising avenue is to connect the present NMF-LAB framework with dynamic classification models. In particular, combining NMF-LAB with the NMF-VAR formulation \citep{satoh2025}, where past data are used as covariates, naturally extends the method toward dynamic logit-type models. Such an approach would allow current class labels to be predicted from past observations, paralleling the framework of dynamic ordered panel logit models (see, e.g., \citealp{honore2025}). This integration could open new applications in time series classification, longitudinal analysis, and sequential decision-making. Overall, this study highlights the potential of NMF-LAB as a versatile tool for modern classification tasks.

\section*{Acknowledgements}

The author thanks the anonymous reviewers for their constructive comments and insightful suggestions, which helped improve the clarity and quality of this paper. 
\section*{Statements and Declarations}

\textbf{Funding.} This work was partly supported by JSPS KAKENHI Grant Numbers 22K11930, 25K15229, 	24K03007,  25H00482 and the project research fund by the Research Center for Sustainability and Environment at Shiga University.
\textbf{Conflicts of Interest.} On behalf of all authors, the corresponding author states that there is no conflict of interest.
\textbf{Ethical Approval.} Not applicable.

\textbf{Data Availability.} The datasets generated or analyzed during the current study are available from the corresponding author on reasonable request.
\textbf{Code Availability.} The R package \texttt{nmfkc} is available at \url{https://github.com/ksatohds/nmfkc}.

\bibliography{ksatoh2025.bib}

\end{document}